\documentclass[journal]{IEEEtran}

\usepackage{times}
\usepackage{epsfig}
\usepackage{graphicx}
\usepackage{amsmath}
\usepackage{amssymb}

\usepackage{graphicx}
\usepackage{amsmath}
\usepackage{amssymb}
\graphicspath{{images/}}
\usepackage{mathtools}
\usepackage{adjustbox}
\usepackage{caption}
\usepackage{url}
\usepackage{subcaption}
\usepackage{array}
\usepackage[usenames,dvipsnames]{color}

\title{\LARGE \bf
GRIP++: Enhanced Graph-based Interaction-aware Trajectory Prediction for Autonomous Driving
}

\author{Xin Li, Xiaowen Ying, Mooi Choo Chuah\\
Department of Computer Science and Engineering, Lehigh University\\
{\tt\small xincoder@gmail.com, xiy517@lehigh.edu, chuah@cse.lehigh.edu}}

\begin{document}

\maketitle
\thispagestyle{empty}
\pagestyle{empty}

\begin{abstract}

Despite the advancement in the technology of autonomous driving cars, the safety of a self-driving car is still a challenging problem that has not been well studied. Motion prediction is one of the core functions of an autonomous driving car. Previously, we propose a novel scheme called GRIP which is designed to predict trajectories for traffic agents around an autonomous car efficiently. GRIP uses a graph to represent the interactions of close objects, applies several graph convolutional blocks to extract features, and subsequently uses an encoder-decoder long short-term memory (LSTM) model to make predictions. Even though our experimental results show that GRIP improves the prediction accuracy of the state-of-the-art solution by 30\%, GRIP still has some limitations. GRIP uses a fixed graph to describe the relationships between different traffic agents and hence may suffer some performance degradations when it is being used in urban traffic scenarios. Hence, in this paper, we describe an improved scheme called GRIP++ where we use both fixed and dynamic graphs for trajectory predictions of different types of traffic agents. Such an improvement can help autonomous driving cars avoid many traffic accidents. Our evaluations using a recently released urban traffic dataset, namely ApolloScape showed that GRIP++ achieves better prediction accuracy than state-of-the-art schemes. GRIP++ ranked \#1 on the leaderboard of the ApolloScape trajectory competition in October 2019. In addition, GRIP++ runs 21.7 times faster than a state-of-the-art scheme, CS-LSTM. Our code will be available at \textcolor{red}{https://github.com/xincoder/GRIP}.

\end{abstract}
\section{\textbf{Introduction}}
\label{sec:intro}

Nowadays, high-quality and affordable cameras are available in many gadgets, e.g., smart-phones, wireless cameras, autonomous vehicles, that humans own these days. Analyzing images/videos captured by these cameras impacts our daily lives. For example, smart-phones have been using face recognition algorithms \cite{shen2014face, patel2016secure, ying2018liveface} to analyze frames captured by front-cameras (RGB or infrared camera) to recognize users, that improves the security and usability of smart-phones. Smart surveillance video systems which can detect and identify suspects \cite{li2017sbgar, fan2018unsupervised, li2017casheirs, deng2018image, li2018rehar, lin2019improving, li2020robust, li2017uav} help law enforcement personnel maintain a safer living environment. Hand gesture recognition algorithms \cite{feng2016depth, rautaray2012real, mosquera2017identifying, zhu2017multimodal} provide a brand new way for human-computer interaction interfaces to be designed. Model decomposition solutions \cite{jaderberg2014Spatial_decomp, Jian2016channel-decomp, lebedev2014CP_decomp, kim2015Tucker_decomp, li2019dac} make deep learning models run much faster on resource-constrained devices.

Recent technology advancement in the fields of computer vision, sensor signal processing, and hardware designing, etc. have enabled autonomous driving technology to go from the ``likely feasible'' to the ``commercially available'' state. However, recent traffic accidents involving autonomous driving cars from Tesla and Uber in 2018 raised people's concern about the safety of self-driving vehicles. Thus, it is extremely important to improve the performance of the intelligent algorithms running on autonomous driving cars. One important example of such intelligent algorithms is the prediction of the future trajectories of the surrounding traffic agents, e.g., vehicles, pedestrians, bicycles, etc. One can avoid traffic accidents if each autonomous driving car involved could precisely predict the locations of its surrounding objects. 

Accurately predicting the motion of surrounding objects is an extremely challenging task, considering that many factors can affect the future trajectory of an object.  Prior works \cite{toledo2009imm, houenou2013vehicle, schreier2014bayesian, tran2014online, schlechtriemen2015will} proposed to predict future locations by recognizing maneuver (change lanes, brake, or keep going, etc.). However, these methods fail to predict the positions of objects accurately when they infer wrongly the type of maneuver. Typically such wrong inference happens when a scheme makes a prediction only based on sensors like GPS that misses visual clues, e.g., turn signals. Then, Karasev et al. \cite{karasev2016intent} proposed to predict the motion of pedestrians by modeling their behaviors as jump-Markov processes. Unfortunately, their proposed method requires a semantic map and knowledge of one or several goals of the pedestrian, which is not useful in the autonomous driving scenario because an autonomous driving car cannot know the destination of a pedestrian (or other objects) in advance. Bhattacharyya et al.  \cite{bhattacharyya2018long} tried to predict the bounding boxes of objects in RGB camera frames by predicting future vehicle odometry sequence. Yet, the predicted bounding boxes in RGB frames still need to be mapped to the coordinate system of the self-driving car to allow the self-driving car to make a correct response to these predicted locations. 

Besides, few of the schemes we discussed above take the states of surrounding objects into account. We argue that the motion states of surrounding objects are crucial for motion prediction especially in the field of autonomous driving. In autonomous driving scenarios, there are different types of nearby traffic agents, e.g., cars, pedestrians, bicycles, buses. These traffic agents have various shapes, dynamics and different movement patterns. To ensure safe operations of autonomous vehicles, their perception and navigation systems should be able to analyze motion patterns of surrounding traffic agents and predict their future locations so that autonomous vehicles can make better driving decisions.

In \cite{itsc19}, we have proposed a robust and efficient object trajectory prediction scheme for autonomous driving cars, namely GRIP, that can infer future locations of nearby objects simultaneously and is trainable end-to-end. Our preliminary results using two large highway datasets show that our approach performs better than existing schemes. However, we did not evaluate our scheme in urban driving environment. Driving in an urban environment is much more challenging than driving on a highway. Urban traffic has more uncertainties and may have more complex road conditions, and diverse traffic agents. Different types of traffic agents have varying motion patterns and their behaviors affect one another. In addition, in \cite{itsc19}, we use a fixed graph to represent the relationship between traffic agents. Such an approach may suffer from performance degradation when it is being used in urban traffic scenarios. Thus, in this paper, we propose an improved scheme called GRIP++ which utilizes both fixed and dynamic graphs to capture the complex interactions between different types of traffic agents for better trajectory prediction accuracy. 

In summary, our contributions of this paper include:
\begin{itemize}
\item An improved object trajectory prediction scheme to precisely predict future locations of various types of traffic agents surrounding an autonomous driving car. 
\item The proposed scheme considers the impact of inter-agent interactions on the motion. 
\item Extensive evaluation using both highway and urban traffic datasets show that our scheme achieves higher accuracy and runs an order of magnitude faster than existing schemes.
\end{itemize}

The rest of this paper is organized as follows. In Section \ref{sec:related_work}, we briefly discuss related work followed by the problem formulation in Section \ref{sec:problem_formulation}. 
In Section \ref{sec:proposed_solution}, we describe our proposed object trajectory prediction scheme and implementation details. We report our experimental results in Section \ref{sec:experiments}. Finally, we conclude this paper in Section \ref{sec:conclusion}.
\section{\textbf{Related work}}
\label{sec:related_work}
\noindent{\bf Conventional Methods on Trajectory Prediction}\\
The problem of trajectory prediction has been extensively studied by researchers over many years. Classical approaches include Monte Carlo Simulation \cite{danielsson07}, Bayesian networks \cite{lefevre11}, Hidden Markov Models (HMM) \cite{firl12} etc. These methods typically focus on analyzing objects based on their previous movements and can only be used in simple traffic scenarios with few interactions among vehicles but such methods may not work well in scenarios involving heterogeneous types of vehicles and pedestrians. Other traditional motion prediction methods are either Markovian maneuver intention estimation-based \cite{schreier2014bayesian, tran2014online} or prototype-trajectory based methods \cite{ferguson15}. Such methods have limitations, e.g., they fail to predict the intent of traffic agents accurately if they infer the wrong type of maneuvers or they are computationally very expensive. The authors in \cite{chen16} combine the two techniques to develop a dictionary learning algorithm called Augmented Semi Non-Negative Sparse Coding (ASNSC). However, ASNSC predicts the intents only based on spatial features while ignoring the environmental context that may influence an object's intent.

Researchers have also attempted to predict trajectories for crowds by modeling pedestrians' behaviors and interactions. For example, \cite{bera16, bera17} combine an Ensemble Kalman Filter and human motion model to predict the trajectories of crowds. Ma et al \cite{ma18} extend such methods to general traffic scenarios where they predict the trajectories of multiple traffic agents by considering kinematic and dynamic constraints. However, they assume perfect sensing, shape and dynamics information for all traffic agents which often are not available in real life. 

\noindent{\bf Recent Deep Learning Based Models for Trajectory Prediction}\\
In recent years, deep learning based methods, e.g., Long Short term Memory (LSTM) based methods, have been proposed for maneuver classification and trajectory prediction, e.g., \cite{khosroshahi17, altche17}. Typically such methods require ideal road conditions, e.g., clear road lanes, no other types of traffic agents or perfect knowledge of surrounding objects. For example, \cite{park18} used one LSTM based encoder to study pattern underlying past trajectory and another LSTM decoder predicts future trajectory but they assume the ego vehicle knows the relative speed and locations of nearby vehicles.  Recently, researchers have realized such limitations and started exploring possible solutions. Thus, we merely summarize the more recent works that take inter-object interactions into account here. 

In \cite{chandra2018}, the authors presented a LSTM-CNN hybrid network called TraPhic for trajectory prediction. Specifically, they take into account heterogeneous interactions that implicitly accounts for the varying dynamics and behaviors of different road agents and use a semi-elliptical region (horizon) in front of each road agent to model horizon-based interactions which implicitly models the driving behavior of each road agent. A LSTM is used to model each road agent. A horizon map is created by pooling together the hidden states of the horizon agents and a neighborhood map is created using hidden states of all agents in the defined neighborhood. Such a scheme is computationally expensive and certain movement information is lost by using CNN to pool the hidden states of nearby agents which limits the accuracy it can achieve. To overcome such limitation, another scheme is proposed in \cite{ma2018trafficpredict} where an instance layer is used to learn instances' movements and interactions and a category layer is used to learn the similarities of instances belonging to the same type of traffic agents to refine the prediction. This scheme performs better than TraPhic but its computation cost remains high since an LSTM is used for each traffic agent in the neighborhood.

Luo et al. proposed a convolutional network for fast object detection, tracking and motion forecasting in \cite{luo2018fast}. Their model takes bird's eye view LiDAR data as input and processes 3D convolutions across space and time. Then, two extra branches of convolutional layers are added: one of them calculates the probability of being a vehicle at a given location and another predicts the bounding box over the current frame as well as several frames in the future. They believe that such a structure can forecast motion because the model can learn velocity and acceleration features from the input of multiple frames. However, the forecasting branch simply takes the 3D convolutional feature map as an input, so visual features of all objects are represented in the same feature map. This results in the model losing track of objects and hence cannot perform well in a scene with crowded objects. 
% However, the forecasting branch only consists of a single set of parameters, so all objects will be predicted using these parameters regardless of the type (pedestrian or vehicle), location or speed of a specific object. Thus, it cannot handle the situation when objects move at varying speeds. 

In addition, Deo et al. \cite{deo2018multi, deo2018would} proposed a unified framework for surrounding vehicles' maneuver classification and motion prediction on freeways. First, an LSTM model is used to represent the track histories and relative positions of all observed cars (the one being predicted and its nearby vehicles) as a context vector. Then, this context  is used for maneuver classification and another LSTM is used to predict the vehicle's future position. Considering that the LSTM model fails to capture the interdependencies of the motion of all cars in the scene, they later enhance their scheme by adding convolutional social pooling layers in \cite{deo2018convolutional}. This improved model has access to the motion states of surrounding objects and their spatial relationships and hence improves the accuracy of future motion prediction. However, all of these models merely predict the trajectory of one specific car (the one in the middle position) each time. Hence, these existing approaches require intensive computation power if they want to predict trajectories of all surrounding objects which is highly inefficient. Besides, these schemes are maneuver based and hence their performance suffer when wrong classifications of the maneuver types occur.

\section{\textbf{Problem Formulation}}
\label{sec:problem_formulation}

Before introducing our proposed scheme, we would like to formulate the trajectory prediction problem as one which estimates the future positions of all objects in a scene based on their trajectory histories. Specifically, the inputs $X$ of our model are trajectory histories (over $t_h$ time steps) of all observed objects: 

\begin{equation}
X = [p^{(1)}, p^{(2)} \cdots, p^{(t_h)}]
\end{equation}
where, 
\begin{equation}
\label{equ:position}
p^{(t)} = [x_0^{(t)}, y_0^{(t)}, x_1^{(t)}, y_1^{(t)}, \cdots, x_n^{(t)}, y_n^{(t)}]
\end{equation}
are the co-ordinates of all observed objects at time $t$, and $n$ is the number of observed objects. This format is the same as what Deo et al. defined in \cite{deo2018multi} and \cite{deo2018convolutional}. Global coordinates are used here. Using relative measurement in the ego-vehicle-based coordinate system will improve the prediction accuracy, but will be left for future work. 

Considering that we feed track histories of all observed objects into the model, we argue that it makes more sense to predict future positions for all of them simultaneously for an autonomous driving car. Thus, instead of only predicting the position of one particular object as being done in \cite{deo2018multi} and \cite{deo2018convolutional}, our proposed model outputs $Y$, predicted future positions of all observed objects from time step $t_h+1$ to $t_h+t_f$:
\begin{equation}
Y = [p^{(t_h+1)}, p^{(t_h+2)}, \cdots, p^{(t_h+t_f)}]
\end{equation}
where $p^{(t)}$ is the same as Eq. (\ref{equ:position}) and $t_f$ is the predicted horizon. 
\section{\textbf{Proposed Scheme}}
\label{sec:proposed_solution}

\begin{figure*}[ht]
	\centering
	\includegraphics[width=0.8\textwidth]{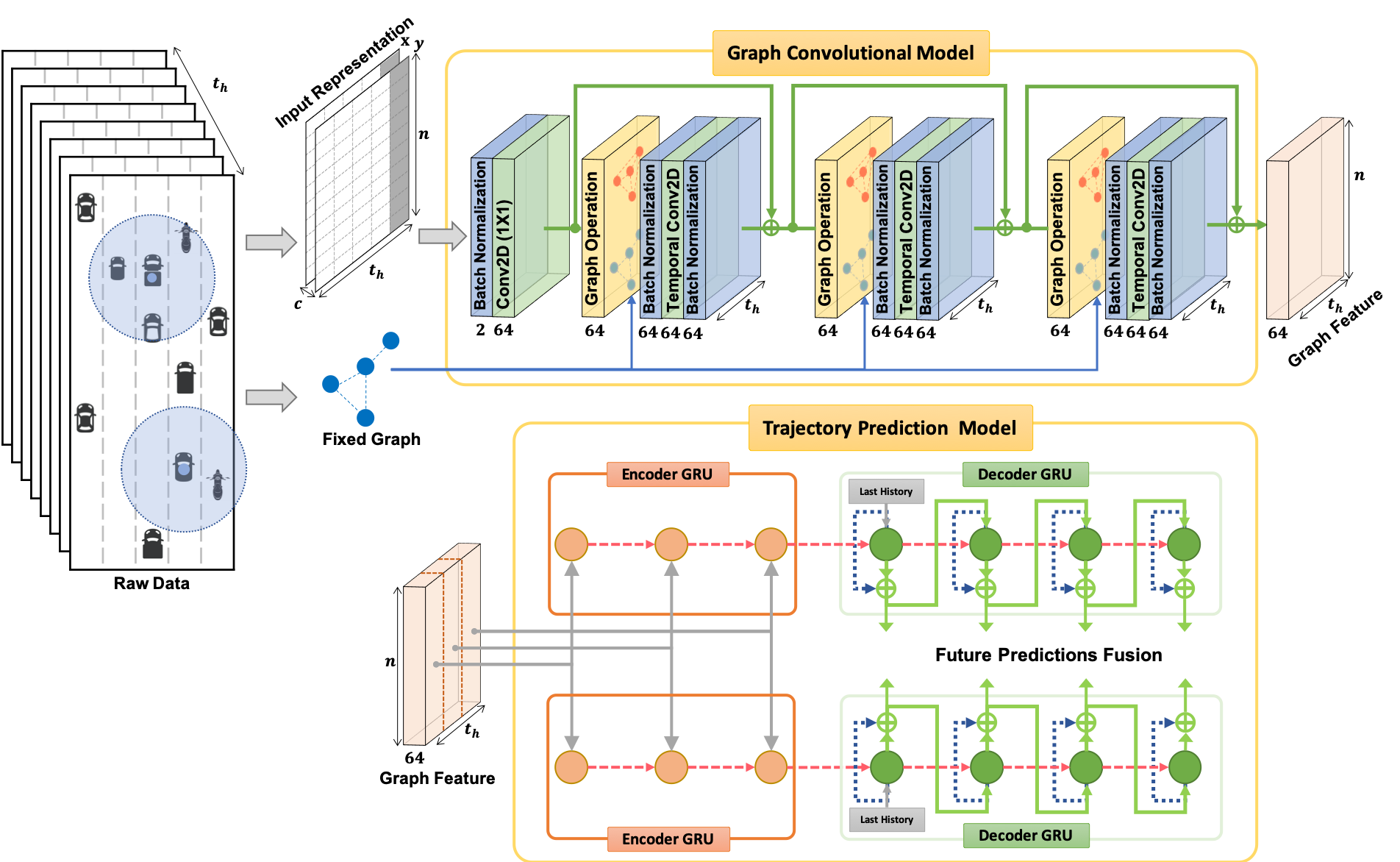}
	%\captionsetup{font=footnotesize}	
	\caption{The architecture of the proposed Scheme.}
	\label{fig:framework}
\end{figure*}

To solve the limitations of existing approaches, we propose GRIP++, a novel deep learning model for object trajectory prediction in this section. Our model, illustrated in Figure \ref{fig:framework}, consists of three components: (1) Input Preprocessing Model, (2) Graph Convolutional Model, and (3) Trajectory Prediction Model. 

%%%%%%%%%%%%%%%%%%%%%%%%%%%%%%%%%%%%%%%%%%%%%%%%
% Input Processing Model
%%%%%%%%%%%%%%%%%%%%%%%%%%%%%%%%%%%%%%%%%%%%%%%%
\subsection{\textbf{Input Preprocessing Model}}

\subsubsection{\textbf{Input Representation}} 
\hfill

Before feeding the trajectory data of objects into our model, we convert the raw data into a specific format for subsequent efficient computation. Assuming that $n$ objects in a traffic scene were observed in the past $t_h$ time steps, we represent such information in a 3D array $F_{input}$ with a size of $(n \times t_h \times c)$ (as shown in Figure \ref{fig:framework}). In this paper, we set $c=2$ to indicate $x$ and $y$ coordinates of an object. Considering that it is easier to predict the velocity of an object than predicting its location, we calculate velocities $(p^{t+1} - p^{t})$ before feeding the data into our model. 

\subsubsection{\textbf{Graph Construction}} 
\label{subsec:graph_construction}
\hfill

Considering that, in the autonomous driving application scenario, the motion of an object is profoundly impacted by the movements of its surrounding objects. This is highly similar to people's behaviors on a social network (one person is usually to be impacted by his/her friends). This inspires us to represent the inter-object interaction using an undirected graph $G = \{V, E\}$ as what researchers have done for a social network. 

In this graph, each node in node set $V$ corresponds to an object in a traffic scene. Considering that each object may have different states at different time steps, the node set $V$ is defined as $V = \{v_{it}|i = 1, \cdots, n, t = 1, \cdots, t_h\}$, where $n$ is the number of observed objects in a scene, and $t_h$ is the observed time steps. The feature vector $v_{it}$ on a node is the coordinate of $i$th object at time $t$.

At each time step $t$, objects that have interactions should be connected with edges. In the autonomous driving application scenario, such an interaction happens when two objects are close to each other. Thus, the edge set $E$ is composed of two parts: (1) The first part describes the interaction information between two objects in spatial space at time $t$. We call it a ``spatial edge'' and denote it as $E_S = \{v_{it}v_{jt}|(i,j \in D)\}$, where $D$ is a set in which objects are close to each other. In this paper, we define that two objects are close if their distance is shorter than a threshold of $D_{close}$. In Figure \ref{fig:framework}, we demonstrate this concept on ``Raw Data" using two blue circles with a radius of $D_{close}$. All objects within the blue circle are regarded as close to the one located in the middle of the circle. Thus, the top object has three close neighbors, and the lower one only has one neighbor. (2) The second part is the inter-frame edges, which represents the historical information frame by frame in temporal space. Each observed object in one time-step is connected to itself in another time-step via the temporal edge and such edges are denoted as $E_F = \{v_{it}v_{i(t+1)}\}$. Thus, all edges in $E_F$ of one particular object represent its trajectory over time steps. 

To make the computation more efficient, we represent this graph using an adjacency matrix $A = \{A_0, A_1\}$, where $A_0$ is an identity matrix $I$ representing self-connections in temporal space, and $A_1$ is a spatial connection adjacency matrix. Thus, at any time $t$,
\begin{equation}
A_0[i][j] (or A_1[i][j]) = \left\{
    \begin{array}{@{} l c @{}}
    1, \text{if edge $\langle v_{it}, v_{jt} \rangle \in E$} \\
    0, \text{otherwise}
    \end{array}\right.
\end{equation}
Both $A_0$ and $A_1$ have a size of $(n \times n)$, where $n$ equals to the number of observed objects in a scene. Such a graph is constructed based on a manually designed rule, so it is fixed once the input data is given and will not change during the training phase. Thus, we called it ``Fixed Graph'' (the blue graph symbol in Figure \ref{fig:framework}) .

\vspace{-3mm}
%%%%%%%%%%%%%%%%%%%%%%%%%%%%%%%%%%%%%%%%%%%%%%%%
% Graph Convolutional Model
%%%%%%%%%%%%%%%%%%%%%%%%%%%%%%%%%%%%%%%%%%%%%%%%
\subsection{\textbf{Graph Convolutional Model}}
Given a preprocessed input data (Input Representation) $F_{input} := \mathbb{R} ^{n \times t_h \times c}$, the Graph Convolutional Model first passes it through a 2D Convolutional layer with $(1 \times 1)$ kernel size (``Conv2D (1x1)'' in Figure \ref{fig:framework}) to increase the number of channel. It maps the 2-dimensional input data (x,y coordinates) into a higher-dimensional space, which helps the model learn a good representation for the trajectory prediction task. Thus, its output has a shape of $(n \times t_h \times C)$, where $C$ is the new number of channel ($C=64$ in Figure \ref{fig:framework}). 

After that, the input data is fed into several graph operations as well as temporal convolutions. These graph operations are designed to handle the inter-object interaction in spatial space, and the temporal convolutions are used to capture useful temporal features, e.g., the motion pattern of one object. Thus, as shown in Figure \ref{fig:framework} (3 Graph Operation layers and 3 Temporal Convolution layers are illustrated), one Temporal Convolution layer is added to the end of each Graph Operation layer in this Graph Convolutional Model to process the input data spatially and temporally alternatively. 

Batch Normalization layers are employed to improve the training stability of our model. Besides, skip connections (green polylines) are used to make sure that the model can propagate larger gradients to initial layers, and these layers also could learn as fast as the final layers.

\subsubsection{\textbf{Graph Operation Layer}} 
A graph operation layer takes the interactions of surrounding objects into account. Each Graph Operation layer consists of two graphs: (i) a Fixed Graph (adjacency matrix $A$ described in the previous section, blue graph symbols in Figure \ref{fig:framework}) constructed based on the current input, and (ii) a trainable graph (denoted as $G_{train}$)(shown in orange graph symbols in the Graph Operation block in Figure \ref{fig:framework}) with the same shape as the Fixed Graph. 

To make sure the value range of feature maps remain unchanged after performing graph operations, we normalize Fixed Graph $A$ using the following equation:

\begin{equation}
G_{fixed}^{j} = \Lambda_j^{-\frac{1}{2}} A_j \Lambda_j^{-\frac{1}{2}}
\end{equation}
where $\Lambda_{j}$ is computed as:

\begin{equation}
\Lambda_{j}^{ii}  = \sum_{k} (A_j^{ik}) + \alpha
\end{equation}
we set $\alpha=0.001$ to avoid empty rows in $A_j$.

Considering that the Fixed Graph $G_{fixed}$ is constructed based on a manually designed rule, it may not be able to represent the interactions of objects properly. In this paper, to solve this problem, we sum the Fixed Graph with the trainable graph, so that the trainable graph can be trained to alleviate the limitation of the Fixed Graph. Thus, once a Graph Operation layer takes an input $f_{conv}$ from its previous layer, the output feature map $f_{graph}$ is calculated as:

\begin{equation}
f_{graph} = \sum_{j=0}^{1}  (G_{fixed}^{j} + G_{train}^{j})f_{conv}
\end{equation}

Graph Operation layers do not change the size of features, so $f_{graph}$ has a size of $(n \times t_h \times C)$.

\subsubsection{\textbf{Temporal Convolutional Layer}} Then, we feed the generated feature $f_{graph} := \mathbb{R} ^{n \times t_h \times C}$ to a Temporal Convolutional layer. We set the kernel size of a Temporal Convolutional layer to $(1 \times 3)$ to force them to process the data along the temporal dimension (second dimension). Appropriate paddings and strides are added to make sure that each layer has an output feature map with the expected size.

\vspace{-3mm}
%%%%%%%%%%%%%%%%%%%%%%%%%%%%%%%%%%%%%%%%%%%%%%%%
% Trajectory Prediction Model
%%%%%%%%%%%%%%%%%%%%%%%%%%%%%%%%%%%%%%%%%%%%%%%%
\subsection{\textbf{Trajectory Prediction Model}}
The Trajectory Prediction Model consists of several networks. These networks share the same Seq2Seq structure but will be trained for different weights. In Figure \ref{fig:framework}, we show two Seq2Seq networks. Each network takes the Graph Feature (generated by the Graph Convolutional Model) as its input. Feature vectors (at each temporal dimension) of the Graph Feature are fed into the corresponding input cell of the Encoder GRU (gray arrows in Figure \ref{fig:framework}). 

Then, the hidden feature of the Encoder GRU, as well as coordinates of objects at the previous time step, are fed into a Decoder GRU to predict the position coordinates at the current time step. Specifically, the input of the first decoding step (gray ``Last History'' boxes in Figure \ref{fig:framework}) is the coordinates of objects at the ``Last History'' step (corresponding to the gray column of the Input Representation in Figure \ref{fig:framework}), and the output of the current step is fed into the next GRU cell. Such a decoding process is repeated several times until the model predicts positions for all expected time steps ($t_f$) in the future. Because few traffic-objects move in a constant velocity, we force the model to predict the change of velocity by adding a residual connection (blue dashed lines in Figure \ref{fig:framework}) between the input and the output to each cell of the Decoder GRU. The impact of using such a residual connection will be discussed in the Experiments section (Section \ref{sec:experiments}).

Finally, once we get the predicted results of these Seq2Seq networks, we average the results (predicted velocities) at each time step. After getting the averaged predicted velocities, we add them $(\Delta x, \Delta y)$ back to the last historical location $(p^{(th)})$ to convert the predicted results to $(x, y)$ coordinates.

The key difference between GRIP++ and GRIP are
\begin{itemize}
\item GRIP++ takes velocity ($\Delta$ x, $\Delta$ y) as input while GRIP takes (x,y) coordinates as input
\item GRIP++ considers both fixed and trainable graphs while GRIP merely considers fixed graphs in the graph convolution submodule.
\item GRIP++ uses 3 blocks in the graph convolution model and adds batch normalization while GRIP uses 10 blocks in the graph convolution model without the batch normalization layers. In addition, GRIP++ uses skip connections.
\item GRIP++ uses GRU networks while GRIP uses LSTM networks. GRIP++ also uses three encoder-decoder blocks for trajectory prediction and average the results while GRIP merely uses a single encoder-decoder block for trajectory prediction.
\end{itemize}

%%%%%%%%%%%%%%%%%%%%%%%%%%%%%%%%%%%%%%%%%%%%%%%%
% Implementation Details
%%%%%%%%%%%%%%%%%%%%%%%%%%%%%%%%%%%%%%%%%%%%%%%%
\subsection{\textbf{Implementation Details}}
\label{sec:implementation_details}
Our scheme is implemented using Python Programming Language and PyTorch Library \cite{paszke2017automatic}. We report the implementation details of our scheme and the settings of important parameters as follows.

\textbf{Input Preprocessing Model:} In this paper, we process a traffic scene within 180 feet ($\pm$ 90 feet). All objects within this region will be observed and predicted in the future. While constructing the graph, we consider two objects are close if their distance is less than 25 feet ($D_{close}=25$).  Thus, any pair of objects within 25 feet are connected using a spatial edge, $e_s \in E_S$. Please refer to our ablation study in section \ref{subsec:ablation_study} for more details.

\textbf{Graph Convolutional Model:} As shown in Figure \ref{fig:framework}, we use a $(1 \times 1)$ convolutional layer to increase the channel of input data to 64. The Graph Convolutional Model consists of 3 Graph Operation layers. Each of these Graph Operation layers is followed by a Temporal Convolution layer. All Temporal Convolution layers have a convolutional kernel with a size of $(1 \times 3)$. We set $stride=1$ and appropriate padding to maintain the shape of feature maps. Thus, the output of the Graph Convolutional Model has a size of $(n \times t_h \times 64)$. To avoid overfitting, we randomly dropout features (0.5 probability) after each graph operation. 
% Each Graph Operation layer and Temporal Convolution layer is followed by a Batch Normalization layer for training stability. 

\textbf{Trajectory Prediction Model:} Both the encoder and decoder of this prediction model are a two-layer GRU (Gated recurrent unit) networks. We set the number of hidden units of these two GRUs equals to $r$ times of the output dimension ($r \times 2 \times n$, where $r$ is used to improve the representation ability,  $n$ is the number of objects and $2$ is the $x, y$ coordinates). In this paper, we choose $r=30$ for its best performance (please refer to Experiments chapter for more discussion). The input of the encoder has 64 channels that are the same as the output of the Graph Convolutional Model.

\textbf{Optimization:} We train our model as a regression task at each time. The overall loss can be computed as: 
%\begin{equation}
%\label{equation: total_loss}
\begin{align}
Loss & = \frac{1}{t_f} \sum\limits_{t=1}^{t_f} loss^{t} \\
		& = \frac{1}{t_f} \sum\limits_{t=1}^{t_f}  \left\Vert Y_{pred}^t - Y_{GT}^t \right\Vert
\end{align}
where $t_f$ is the time step in the future (in Figure \ref{fig:framework}, $t_f = 4$), $loss^t$ is the loss at time $t$, $Y_{pred}$  and $Y_{GT}$ are predicted positions and ground truth respectively. The model is trained to minimize the $Loss$.

\textbf{Training Process:} We train the model using Adam optimizer with default settings in Pytorch Library. We set $batch\_size = 64$ during training.

\section{\textbf{Experiments}}
\label{sec:experiments}
We run our scheme on a desktop running Ubuntu 16.04 with 4.0GHz Intel Core i7 CPU, 32GB Memory, and a NVIDIA Titan Xp Graphics Card.
\vspace{-3mm}
\subsection{\textbf{Datasets}}
We evaluate our scheme on three well known trajectory prediction datasets: NGSIM I-80 \cite{colyar2007us80}, US-101 \cite{colyar2007us101}, and ApolloScape Trajectory dataset \cite{ma2018trafficpredict}.

\textbf{NGSIM Datasets:} Both NGSIM I-80 and US-101 datasets were captured at 10 Hz over 45 minutes and segmented into 15 minutes of mild, moderate and congested traffic conditions. These two datasets consist of trajectories of vehicles on real freeway traffic. Coordinates of cars in a local coordinate system are provided. 

We follow Deo et al. \cite{deo2018multi, deo2018would, deo2018convolutional} to split these two datasets into training and testing sets. One-fourth of the data from each of the three subsets (mild, moderate, and congested traffic conditions) are selected for testing. Each trajectory is segmented into 8 seconds clips that the first 3 seconds are used as observed track history and the remaining 5 seconds are the prediction ground truth. To make a fair comparison, we also do the same downsampling for each segment by a factor 2 as Deo et al. did, i.e. 5 frames per second. The code for dataset segmentation can be downloaded from their Github \footnote{https://github.com/nachiket92/conv-social-pooling}.

\begin{figure}[t]
	\centering
	\begin{subfigure}{0.5 \textwidth}
		\label{fig:uts1}
		\centering
		\includegraphics[width=0.9 \linewidth]{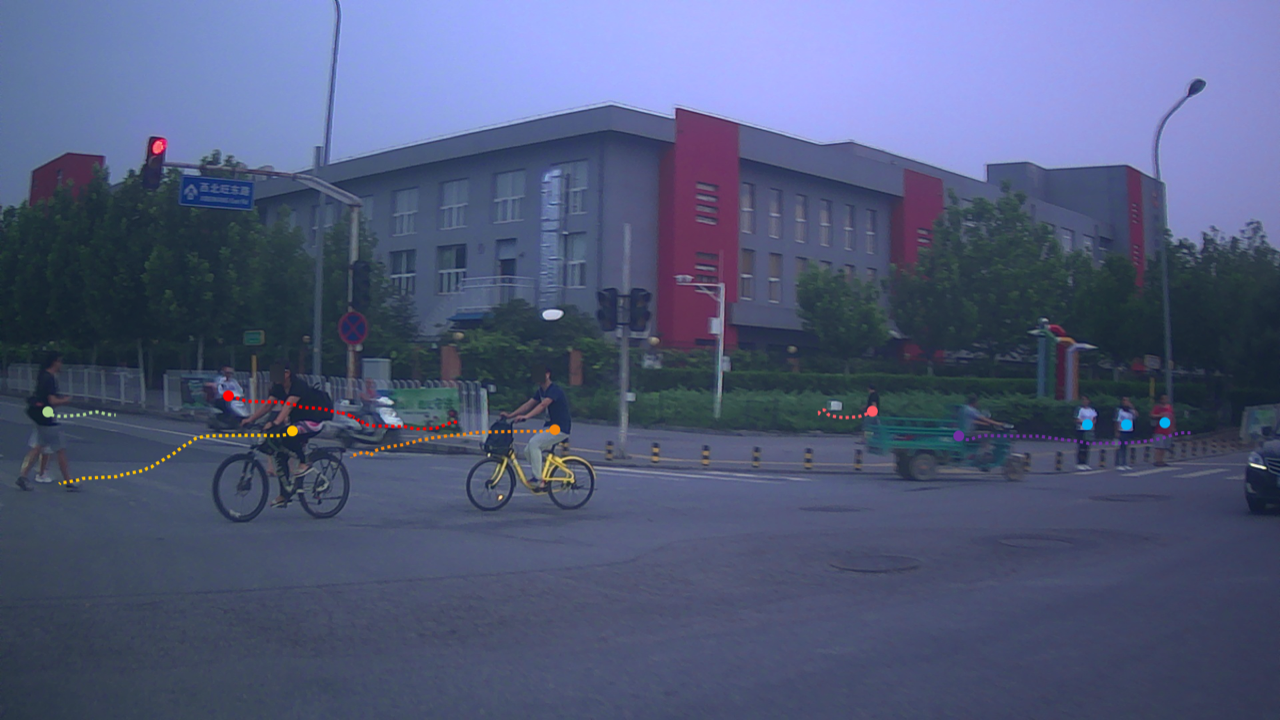}
		\caption{Urban Traffic Scene 1}
		\end{subfigure}
	\begin{subfigure}{0.5 \textwidth}
		\label{fig:uts2}
		\centering
		\includegraphics[width=0.9 \linewidth]{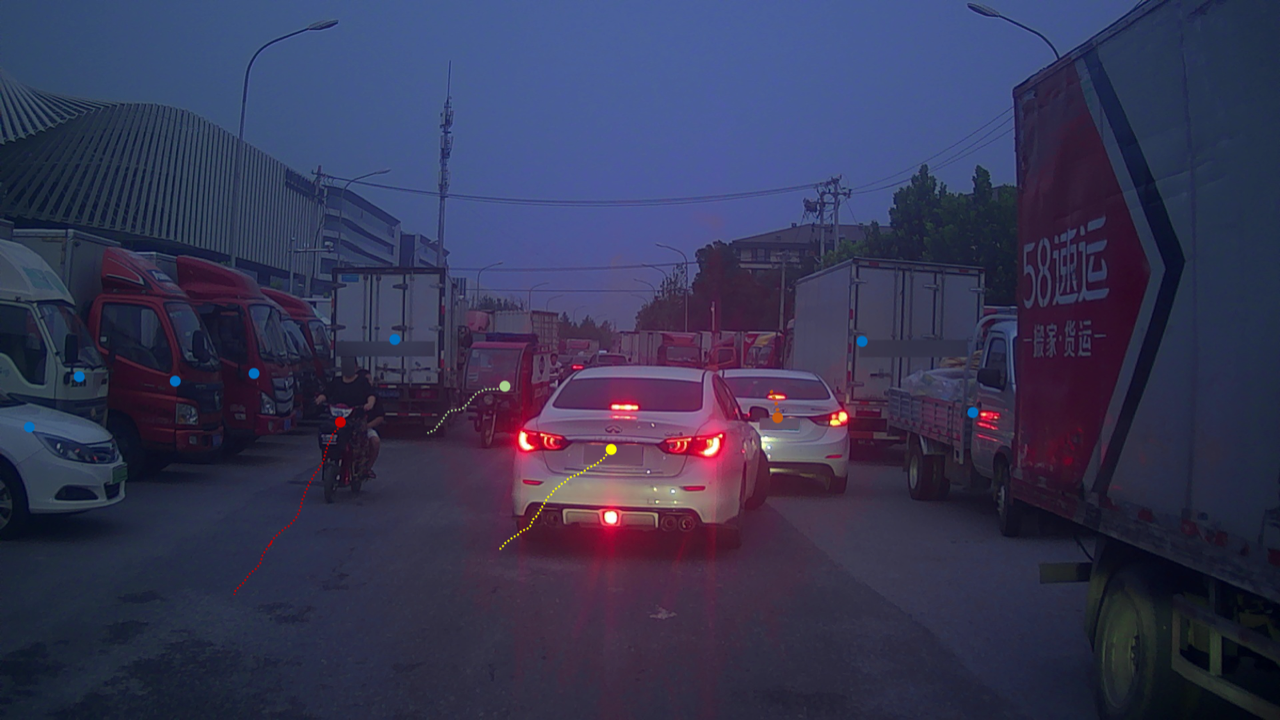}
	\caption{Urban Traffic Scene 2.}
	\end{subfigure}
		%\vspace{-5mm}
		\caption{ApolloScape Urban Traffic Scenes \cite{apolloscape18}}
	\label{fig:urban}
\end{figure}

\textbf{ApolloScape Trajectory Dataset:} The ApolloScape Trajectory dataset is collected by running the Apollo acquisition car \cite{apolloscape18} in urban areas during rush hours. Traffic data, including camera-based images and LiDAR-based point clouds, are collected, and object trajectories are calculated using object detection and tracking algorithms. In total, the trajectory dataset consists of 53min training sequences and 50min testing sequences captured at 2 frames per second. Object id, object type, object position, object size, and heading angle, etc. are provided. Because the data is collected in urban areas, there are five different object types involved: small vehicles, big vehicles, pedestrians, motorcyclists and bicyclists, and others. This particular dataset allows researchers to stress test the trajectory prediction scheme they design for having various types of traffic agents with different behaviors create additional challenges in the design. In Fig~\ref{fig:urban}, we highlight some of these challenges that any object trajectory prediction scheme faces in urban traffic scenarios. In Fig~\ref{fig:urban} (a), we have  traffic agents moving at various speeds in different directions while in Fig~\ref{fig:urban} (b), we have three traffic agents of different sizes with one traffic agent trying to squeeze through a tight space.

During the training phase, we choose 20\% sequences from the training subset for validation and train our model using the remaining 80\% sequences. Once the model is trained, we generate predictions on testing sequences and submit the results to the ApolloScape website for evaluation.

\subsection{\textbf{Metrics}}
\textbf{RMSE:} For NGSIM I-80 and US-101 dataset, we use the same experimental settings and evaluation metrics as \cite{deo2018convolutional} and \cite{kuefler2017imitating}. In this paper, we report our results in terms of the root of the mean squared error (RMSE) of the predicted trajectories in the future (5 seconds horizons). The RMSE at time $t$ can be computed as follows:

\begin{equation}
RMSE = \sqrt{\frac{1}{n} \sum_{i=1}^{n} (Y_{pred}^{t} [i] - Y_{GT}^{t} [i]) ^ 2 } 
\end{equation}
where $n$ is the number of observed (predicted) objects, $Y_{pred}^t$ and $Y_{GT}^t$ are predicted results and ground truth at time $t$ correspondingly. 

\textbf{WSADE} and \textbf{WSFDE:} For ApolloScape Trajectory dataset, we use both Weighted Sum of Average Displacement Error (WSADE) and Weighted Sum of Final Displacement Error (WSFDE) metrics to evaluate the performance. As described on the ApolloScape website, the Average displacement error (ADE) measures the mean Euclidean distance over all the predicted positions and ground truth positions during the prediction time, and the Final displacement error (FDE) is the mean Euclidean distance between the final predicted positions and the corresponding ground truth locations. Because the trajectories of cars, bicyclist and pedestrians have different scales, they use the following weighted sum of ADE (WSADE) and weighted sum of FDE (WSFDE) as metrics.
\begin{equation}
WSADE = D_v \cdot ADE_v + D_p \cdot ADE_p + D_b \cdot ADE_b
\end{equation}
\begin{equation}
WSFDE = D_v \cdot FDE_v + D_p \cdot FDE_p + D_b \cdot FDE_b
\end{equation}
where $D_v$, $D_p$, and $D_b$ are related to reciprocals of the average velocity of vehicles, pedestrian and cyclist in the dataset. They set the values of these three variables to 0.20, 0.58, 0.22 respectively.

\subsection{\textbf{Ablation Study}}
\label{subsec:ablation_study}
\begin{figure}[h]
	\centering
	\includegraphics[width=0.48\textwidth]{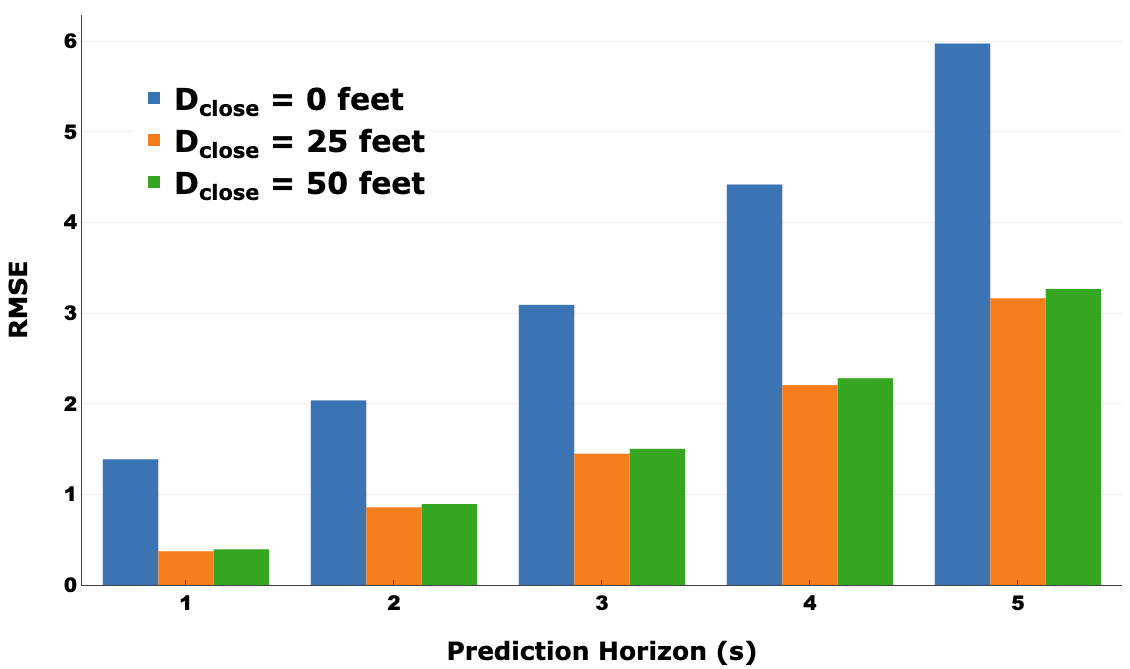}
	\caption{Comparison among various $D_{close}$ values.}
	\label{fig:Dclose}
	\vspace{-4mm}
\end{figure}

In this subsection, we conduct two ablation studies:

(1) We defined a threshold $D_{close}$ in section \ref{subsec:graph_construction}. Two objects within $D_{close}$ range are regarded as close to each other. We first explore how this threshold impacts the performance of our model. In Figure \ref{fig:Dclose}, we compare results when $D_{close}$ is set to different values. One can see that the prediction error when $D_{close}=0$ (when none of the surrounding objects are considered, blue bars in Figure \ref{fig:Dclose}) is higher than the results when $D_{close}>0$ (taking nearby objects into account). Thus, considering the surrounding object indeed helps our model make a better prediction.

\newcolumntype{S}{>{\centering\arraybackslash}m{0.05\textwidth}}
\newcolumntype{L}{>{\centering\arraybackslash}m{0.08\textwidth}}
\newcolumntype{X}{>{\centering\arraybackslash}m{0.10\textwidth}}
\newcolumntype{M}{>{\centering\arraybackslash}m{0.14\textwidth}}
\begin{table*}[t!]
  \centering  
  \caption{Root Mean Square Error (RMSE) for trajectory prediction on NGSIM I-80 and US-101 datasets. Data are converted into the meter unit. All results except ours are extracted from \cite{deo2018convolutional}. The smaller the value, the better. }
  \begin{adjustbox}{width=\textwidth}
  \begin{tabular}{|L|c c L L X L|S|M|} 
    \hline
    Prediction Horizon (s)& CV & V-LSTM & C-VGMM + VIM \cite{deo2018would} & GAIL-GRU \cite{kuefler2017imitating} & CS-LSTM(M) \cite{deo2018convolutional} & CS-LSTM \cite{deo2018convolutional} & GRIP \cite{itsc19} & GRIP++ ($\bigtriangleup$CS-LSTM) \\
    \hline
    1 & 0.73 & 0.68 & 0.66 & 0.69 & 0.62 & 0.61 & \textbf{0.37} & 0.38 (38\%$\uparrow$ -0.23) \\
    2 & 1.78 & 1.65 & 1.56 & 1.51 & 1.29 & 1.27 & \textbf{0.86} & 0.89 (30\%$\uparrow$ -0.38) \\
    3 & 3.13 & 2.91 & 2.75 & 2.55 & 2.13 & 2.09 & 1.45 & \textbf{1.45} (31\%$\uparrow$ -0.64) \\
    4 & 4.78 & 4.46 & 4.24 & 3.65 & 3.20 & 3.10 & 2.21 & \textbf{2.14} (31\%$\uparrow$ -0.96) \\
    5 & 6.68 & 6.27 & 5.99 & 4.71 & 4.52 & 4.37 & 3.16 & \textbf{2.94} (33\%$\uparrow$ -1.43) \\
    \hline
  \end{tabular}
  \end{adjustbox}
  \label{table:comparison_rmse}
\end{table*}

Also, we notice that the prediction error increases when $D_{close}$ increases from 25 feet (orange bars) to 50 feet (green bars). This is because more objects are used to predict the motion of an object with larger $D{close}$. In real life, a traffic agent is more likely to be only impacted by its closest objects. Thus, considering too many surrounding objects does not help to improve the prediction accuracy. Based on this observation, in this paper, we set $D_{close}=25$ feet as our default setting unless specified otherwise.

\begin{figure}[h]
	\vspace{-4mm}
	\centering
	\includegraphics[width=0.48\textwidth]{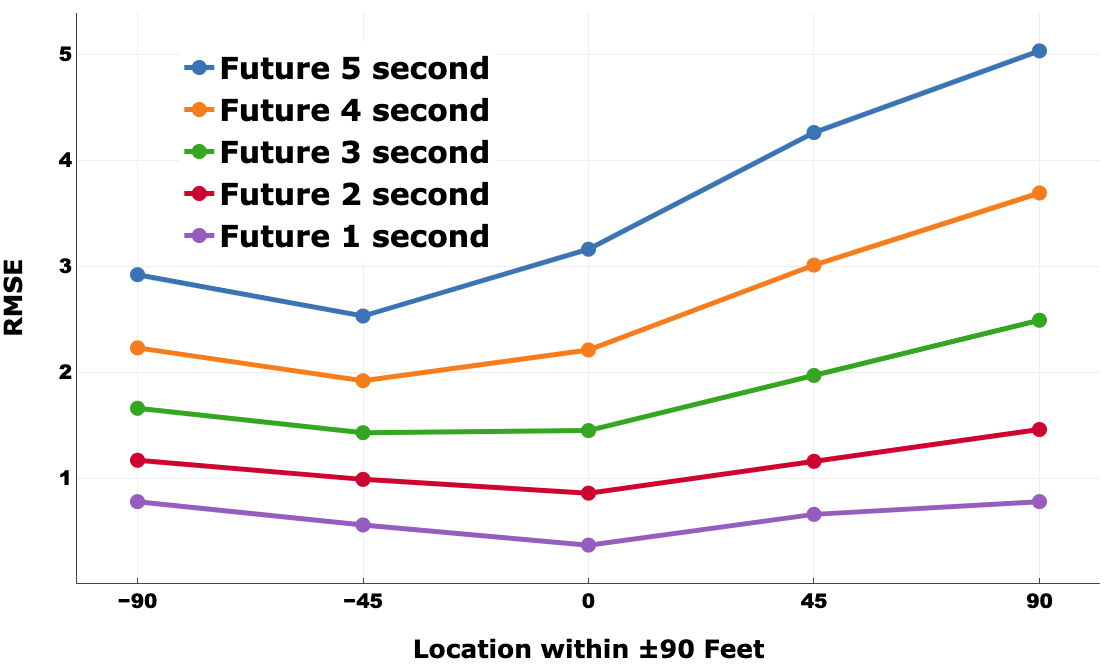}
	\caption{Prediction error at different locations.}
	\label{fig:Location}
	\vspace{-4mm}
\end{figure}

(2) Given an input stream consisting of observed objects' past trajectories, our model is able to predict future trajectories for all observed objects.  Thus, in Figure \ref{fig:Location}, we report the prediction error for objects at different locations, e.g., $-60$ or $-45$ feet,  within the observed area. In Figure \ref{fig:Location}, traffic agents are moving from location $-90$ to location $90$ (left to right). 

First, one may notice that the prediction error decreases from location $-90$ to $-45$, and then increases after $-45$. Such an observation is obvious on the top 3 curves (``Future 5/4/3 second''). This is impacted by the clue information from surrounding objects. Because objects are moving from left to right in Figure \ref{fig:Location}, so objects located at $90$ can only observe objects behind them, while objects at $-90$ can only see objects in front of them. Thus, prediction error at $-90$ is lower than the error at $90$ concludes that front objects are more important than behind objects for our trajectory prediction model. This is also the reason why prediction error increases after $-45$ (less and less front objects are observed from left to right). 

In addition, considering that predicting the motion of an object in far future is difficult. Thus, in Figure \ref{fig:Location}, the error of a long time prediction (``Future 5 second'') is higher than a shorter time prediction (``Future 1 second''). 
%Such an observation is consistent with our intuition. 

%\subsection{\textbf{Comparison Results}}
\subsection{\textbf{Experiments on the NGSIM Datasets}}
In this subsection, we compare our proposed scheme to the following baselines (as done in \cite{deo2018convolutional}) and some existing solutions using NGSIM datasets:
\begin{itemize}
\item Constant Velocity (CV): This is a baseline that only uses a constant velocity Kalman filter to predict trajectories in the future. 
\item Vanilla LSTM (V-LSTM): A baseline that feeds a tack history of the predicted object to an LSTM model to predict a distribution of its future position. 
\item C-VGMM + VIM: In \cite{deo2018would}, Deo et al. propose a maneuver based variational Gaussian mixture model with a Markov random field based vehicle interaction module. 
\item GAIL-GRU: Kuefler et al. \cite{kuefler2017imitating} use a generative adversarial imitation learning model for vehicle trajectory prediction. However, they use ground truth data for surrounding vehicles as input during prediction phase. 
\item CS-LSTM (M): This is the model that an LSTM model with convolutional social pooling layers proposed by Deo et al. in \cite{deo2018convolutional}. A maneuver classier is included. 
\item CS-LSTM: A CS-LSTM model without the maneuver classifier described in \cite{deo2018convolutional}.
\end{itemize}

Comparison results are reported in Table \ref{table:comparison_rmse}. Our model can predict the trajectories for all observed objects simultaneously, while other schemes listed in Table \ref{table:comparison_rmse} only predict one specific object (in the middle position) each time. Thus, to make a fair comparison, we compute the RMSE for the same objects as other schemes and report the result in the last column, ``GRIP++ ($\bigtriangleup$CS-LSTM)'', of Table \ref{table:comparison_rmse}. Compared to the existing state-of-the-art result (CS-LSTM \cite{deo2018convolutional}), our proposed GRIP++ improves the prediction performance by at least 30\%. One may notice that, after 3 seconds in the future, the prediction error of GRIP++ is a half meter (or longer) shorter than CS-LSTM \cite{deo2018convolutional}. We believe that such an improvement can help an autonomous driving car avoid many traffic accidents.  

Then, compared the result of CS-LSTM(M) to CS-LSTM, one can see that CS-LSTM makes slightly better prediction than CS-LSTM(M). This is consistent with our argument mentioned in Section \ref{sec:related_work} that a wrong classification of maneuver type has an adverse effect on the trajectory prediction. 

Besides, compared to our previous work (GRIP, the second column on the right side in Table \ref{table:comparison_rmse}), GRIP++ achieves comparable results in short prediction (the first three seconds) and better results in the long forecast (at 4 second and 5 second). It proves that our GRIP++ has better capability to extract useful features from historical trajectories and then make a long prediction. The NGSIM datasets only consist of trajectories of vehicles on freeway traffic, which means the motion patterns are similar and straightforward. Thus, our GRIP and GRIP++ have a similar performance on the NGSIM datasets (especially for short-term prediction). However, predicting trajectories in urban scenarios is much more complicated and difficult than in highway scenarios. Thus, in the next chapter, we evaluate the performance of our proposed scheme using a dataset collected in urban areas.

%Besides, we also report RMSE results for all predicted objects in the last column, ``GRIP (ALL)'', of Table \ref{table:comparison_rmse}. It is worth highlighting that:
%\begin{itemize}
%\item All schemes in Table 1 take the same data (an object in the middle position and its surrounding objects) as their inputs. Our model predicts all observed objects simultaneously, while others only predict the one in the central location. 
%\item As we discussed the ablation study subsection \ref{subsec:ablation_study}, objects located at the edge of the observed area, e.g., located at $\pm 90$ feet position, do not have enough surrounding objects as input. Thus, the prediction errors of these objects are high, which results in the results in the column of ``GRIP (ALL)'' are higher than ``GRIP''.
%\end{itemize}
%Even so, our proposed GRIP still achieves better prediction results than all of the other existing solutions. 

\subsection{\textbf{Experiments on the ApolloScape Trajectory Datasets}}
In Table \ref{table:baidu_competition}, we compare our proposed scheme to other methods on ApolloScape leaderboard that have publications. It is obvious that GRIP++ achieves much better prediction results than the TrafficPredict \cite{ma2019trafficpredict} (85\% improvement). StarNet \cite{zhu2019starnet} ranked \#1 in the CVPR2019 trajectory prediction challenge. From Table \ref{table:baidu_competition}, one can see that GRIP++ achieves better performance than StartNet in terms of all metrics (both ADEs and FDEs). 

\begin{table*}[h]
  \centering  
  \caption{Competition Results On ApolloScape Trajectory Dataset.}
  \begin{adjustbox}{width=\textwidth}
  \begin{tabular}{|c|c|c c c |c| c c c|} 
  	\hline
  	Method & WSADE & ADEv & ADEp & ADEb & WSFDE & FDEv & FDEp & FDEb \\
	\hline
	TrafficPredict \cite{ma2019trafficpredict} & 8.5881 & 7.9467 & 7.1811 & 12.8805 & 24.2262 & 12.7757 & 11.1210 & 22.7912 \\
	StarNet \cite{zhu2019starnet} & 1.3425 & 2.3860 & 0.7854 & 1.8628 & 2.4984 & 4.2857 & 1.5156 & 3.4645 \\
	\hline
	\textbf{GRIP++} (Ours) & \textbf{1.2588} & 2.2400 & 0.7142 & 1.8024 & \textbf{2.3631} & 4.0762 & 1.3732 & 3.4155 \\
	\hline
  \end{tabular}
  \end{adjustbox}
  \label{table:baidu_competition}
\end{table*}

\begin{table*}[t!]
  \centering  
  \caption{Changes in performance (in term os WSADE) while adjusting the model. Each time, we only change one setting, and the change is highlighted with a underlined bold font. The smaller the value, the better.}
  \begin{adjustbox}{width=\textwidth}
  \begin{tabular}{|c| c c c c c c c c c | c |} 
    \hline
    Index & BatchNorm & Input & RNN Type & GCN\# & RNN In+Out & GCN Graph & RNN\# & RNN Size (r) & Data Aug. & WSADE\\
     \hline
	B1 (GRIP) & N & Norm & LSTM & 10 & N & FixedOnly & 1 & 2 & N & 7.2352\\
	B2 & \underline{\textbf{Y}} & Norm & LSTM & 10 & N & Fixed Only & 1 & 2 & N & 6.9971\\
	B3 & Y & \underline{\textbf{Velocity}} & LSTM & 10 & N & Fixed Only & 1 & 2 & N & 2.6679\\
	B4 & Y & Velocity & \underline{\textbf{GRU}} & 10 & N & Fixed Only & 1 & 2 & N & 2.0743\\
	B5 & Y & Velocity & GRU & \underline{\textbf{3}} & N & Fixed Only & 1 & 2 & N & 2.0034\\
	B6 & Y & Velocity & GRU & 3 & \underline{\textbf{Y}} & Fixed Only & 1 & 2 & N & 1.5207\\
	B7 & Y & Velocity & GRU & 3 & Y & \underline{\textbf{Fixed + Train}} & 1 & 2 & N & 1.4839\\
	B8 & Y & Velocity & GRU & 3 & Y & Fixed + Train & \underline{\textbf{3}} & 2 & N & 1.3936\\
	B9 & Y & Velocity & GRU & 3 & Y & Fixed + Train & 3 & \underline{\textbf{4}} & N & 1.3863\\
	B10 & Y & Velocity & GRU & 3 & Y & Fixed + Train & 3 & \underline{\textbf{10}} & N & 1.3245\\
	B11 & Y & Velocity & GRU & 3 & Y & Fixed + Train & 3 & \underline{\textbf{40}} & N & 1.3227\\
	B12 & Y & Velocity & GRU & 3 & Y & Fixed + Train & 3 & \underline{\textbf{30}} & N & 1.2803\\
	B13 (GRIP++) & Y & Velocity & GRU & 3 & Y & Fixed + Train & 3 & 30 & \underline{\textbf{Y}} & 1.2588\\
    \hline
  \end{tabular}
  \end{adjustbox}
  \label{table:baidu_competition_changes}
\end{table*}

To understand the reason why our proposed GRIP++ works better than others, we report the changes in performance while adjusting the model (structure and parameters) in Table \ref{table:baidu_competition_changes}. The adjustments we made are listed as follows:

\begin{itemize}
\item \textbf{BatchNorm:} As shown in Figure \ref{fig:framework}, we add a Batch Normalization layer after each Graph Operation layer and Temporal Convolution layer. In Table \ref{table:baidu_competition_changes}, ``\textbf{Y}'' indicates Batch Normalization layers included, and ``\textbf{N}'' indicates excluded. 

\item \textbf{Input:} Two types of input processings are tested. (1) Use the position coordinates as the input of our model while normalizing them to the range of [-1, 1] by dividing all data by the maximum value in the training set (mark as ``\textbf{Norm}'' in Table \ref{table:baidu_competition_changes}). (2) Take the velocity of objects as the input of our model (mark as ``\textbf{Velocity}'' in Table \ref{table:baidu_competition_changes}).

\item \textbf{RNN Type:} There are different types of RNN networks. In this paper, we tried Long short-term memory (LSTM) and Gated Recurrent Units (GRU).

\item \textbf{GCN\#:} We also tried different numbers of GCN layers. 10 means 10 Graph Operation layers and 10 Temporal Convolution layers are used.  

\item \textbf{RNN In+Out:} We argue that it is easier to predict velocity of an object than its location, and add a residual connection between the input and the output to the Decoder GRU. Thus, in Table \ref{table:baidu_competition_changes}, ``\textbf{Y}'' indicates residual connections are used, and ``\textbf{N}'' indicates no residual connections in Decoder GRU. 

\item \textbf{GCN Graph:} In each Graph Operation layer, we add the Fixed graph and a trainable graph before performing the graph operation. We evaluate the effectiveness of the trainable graph.  

\item \textbf{RNN\#:} The number of Seq2Seq networks in the Trajectory Prediction Model is also explored. 3 indicates 3 Seq2Seq networks with the same structure are used, and their results will be averaged. 

\item \textbf{RNN Size (r):} In Section \ref{sec:implementation_details}, we set the hidden size of RNN networks to be $r$ times of the output dimension. Thus, we explore the impact of using different values of $r$. 

\item \textbf{Data Aug.:} To train a model with better generalization capability, we applied data augmentation on the input data, e.g., randomly rotate the input data or enhance the model using testing observed data. ``\textbf{Y}'' indicates data augmentation is applied, and ``\textbf{N}'' indicates no data augmentation. 

\end{itemize}

From Table \ref{table:baidu_competition_changes}, we observe the following conclusions:
\begin{itemize}
\item 1. Comparing B3 to B2, one can see a significant improvement by using velocity as input instead of normalized positions. This verified our argument that predicting the velocity of an object is easier than predicting its location. Predicting the physical position of an object is hard because the position value (x, y coordinates) can be any value, so it has a very large norm that the model cannot easily learn a good weight to handle it. However, the velocity of an object is more constant, no matter where the object locates. 

\item 2. From B5 to B6, another big improvement achieves due to the residual connection between the input and the output of the Decoder GRU. This result proves that the residual connection indeed helps the model learn to adjust the velocity. After adding the residual connection, the model just needs to learn the change of velocity (acceleration). Because the change of velocity (acceleration) is more constant than the velocity itself, predicting acceleration is an easier task for the model to learn, which results in a better performance. 

\item 3. Compared to B6, B7 includes a trainable graph in each Graph Operation layer. B7 makes a better prediction than B6 proves that trainable graphs are indeed trained to cover the shortage of Fixed Graphs. 

\item 4. At B4, we change LSTM to GRU. The scale of improvement is surprising. Usually, GRUs train faster and perform better than LSTMs on less training data. If this experience holds in this task, it means that the amount of data in the ApolloScape dataset is not enough for the LSTM model we use. 

\item 5. Then, at B5, we reduce the number of GCN layers (both Graph Operation layers and Temporal Convolution Layers) from 10 to 3. The simplified model achieves a similar performance. Thus, we use the model with fewer layers for faster speed (training and testing speed). 

\item 6. From B8 to B12, different hidden sizes in our Seq2Seq networks are explored. One can see that the performance initially improves as r increases until $r=30$). After that, the performance degrades (B11 performs worse than B12). With increasing r, we have more parameters to train. The increasing number of parameters helps to capture more information but it will also eventually cause an overfitting problem.

\item 7. Besides the above insights, we also notice that adding Batch Normalization layers, using more Seq2Seq models, and doing data augmentation helps a little bit but not too much in the performance of the model. 

\end{itemize}

%Considering the limitation of the page, we do not discuss all findings but some critical factors in Table \ref{table:baidu_competition_changes}.

Compared B13 to B1 in Table \ref{table:baidu_competition_changes}, one can see that GRIP++ achieves much better prediction results (83\% improvement) on the ApolloScape Trajectory Datasets. As we mentioned above, predicting trajectories in urban scenarios is difficult. This result proves the proposed GRIP++ is more robust and useful in real-world scenarios. 

\subsection{\textbf{Computation Time}}
Computation efficiency is one of the important performance indicators of an algorithm for autonomous driving cars. Thus, we evaluate the computation time of our proposed GRIP and GRIP++, and report the results in Table \ref{table:computation_time}.

To make a fair comparison, we downloaded the code of CS-LSTM \cite{deo2018convolutional} \footnote{https://github.com/nachiket92/conv-social-pooling} and ran it on our machine to collect its computation time. CS-LSTM, GRIP and GRIP++ are implemented using PyTorch. 

\newcolumntype{N}{>{\centering\arraybackslash}m{0.06\textwidth}}
%\newcolumntype{L}{>{\centering\arraybackslash}m{0.08\textwidth}}
\begin{table}[h!]
  \centering  
  \caption{Computation time }
  \begin{adjustbox}{width=0.5 \textwidth}
  \begin{tabular}{|c|c|L|N|} 
    \hline
    Scheme & Predicted \# & Time (s) 128 batch & Time (s) 1 batch\\
    \hline
	CS-LSTM \cite{deo2018convolutional} & 1000 & 0.29 & 35.13\\
	\hline
	GRIP \cite{itsc19} & 1000 & 0.05 & 6.33 \\
    \hline
    GRIP++  & 1000 & \textbf{0.02} & \textbf{1.62} \\
    \hline
  \end{tabular}
  \end{adjustbox}
  \label{table:computation_time}
\end{table}

From Table \ref{table:computation_time}, one can see that, when using 128 batch size, CS-LSTM \cite{deo2018convolutional} needs 0.29s to predict trajectories for 1000 objects, GRIP takes 0.05s (5.8x faster), while our proposed GRIP++ only takes 0.02s (14.5x faster than CS-LSTM). In the autonomous driving application scenario, considering the limited resources, we can only set $batch\_size=1$, so we report the results in the last column of Table \ref{table:computation_time}. It shows that GRIP runs 5.5 times faster and GRIP++ runs 21.7 times faster than CS-LSTM \cite{deo2018convolutional}.

The primary reason that GRIP and GRIP++ run faster than CS-LSTM is that both GRIP and GRIP++ predict trajectories for all observed objects simultaneously, while CS-LSTM only predicts for one object. Besides, GRIP++ consists of much fewer layers than GRIP, which results in its faster speed. We only use 3 Graph Operation layers and 3 Temporal Convolution layers in GRIP++, but GRIP has 10 Graph Operation layers and 10 Temporal Convolution layers. 

\subsection{\textbf{Visualization of Prediction Results}}

\begin{figure*}[ht]
    \centering
    \begin{subfigure}[b]{0.19\textwidth}
        \includegraphics[width=\linewidth]{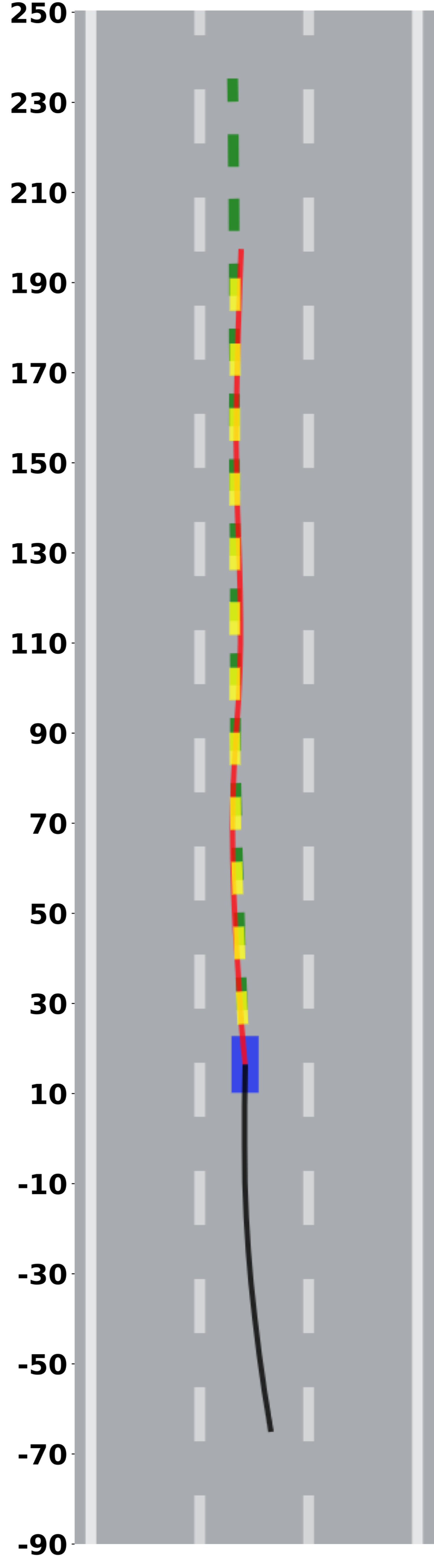}
        \caption{}
        \label{fig:vis_sub_a}
    \end{subfigure}
    \begin{subfigure}[b]{0.19\textwidth}
        \includegraphics[width=\linewidth]{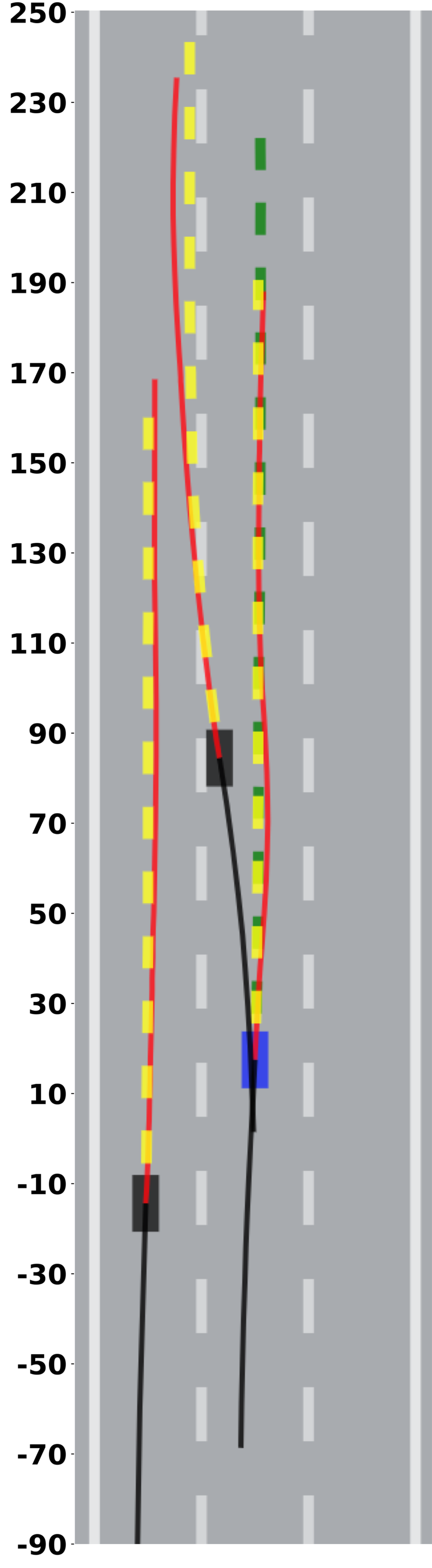}
        \caption{}
        \label{fig:vis_sub_b}
    \end{subfigure}  
    \begin{subfigure}[b]{0.19\textwidth}
        \includegraphics[width=\linewidth]{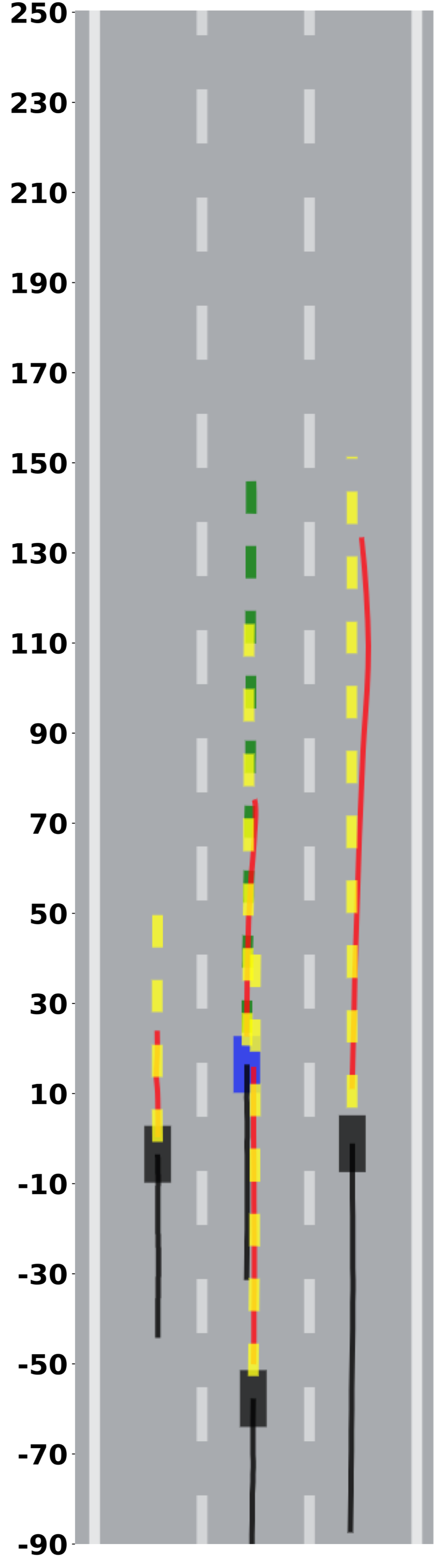}
		 \caption{}       
        \label{fig:vis_sub_c}
    \end{subfigure}
    \begin{subfigure}[b]{0.19\textwidth}
        \includegraphics[width=\linewidth]{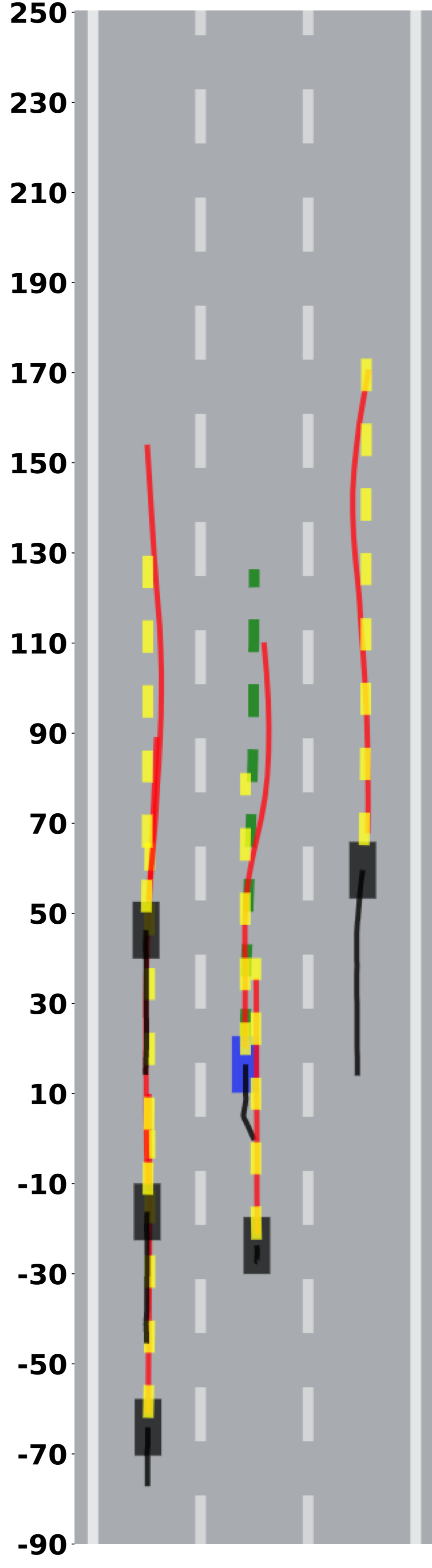}
        \caption{}
        \label{fig:vis_sub_d}
    \end{subfigure}
    \begin{subfigure}[b]{0.19\textwidth}
        \includegraphics[width=\linewidth]{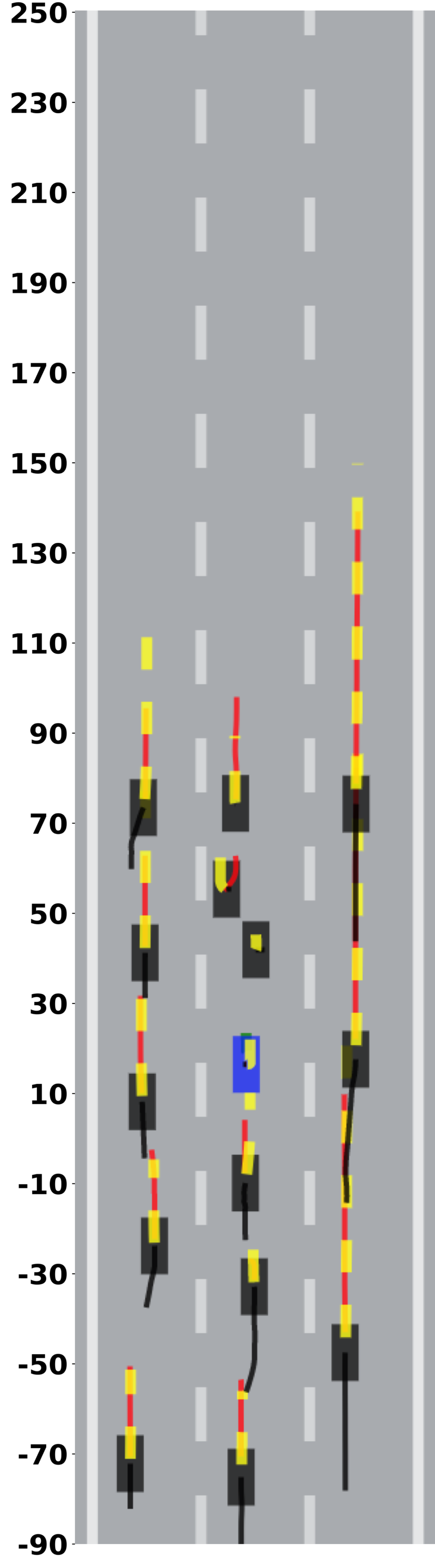}
        \caption{}
        \label{fig:vis_sub_e}
    \end{subfigure}
    \caption{Visualized Prediction Results. \underline{Blue rectangles} are the cars located in the middle which is the car that CS-LSTM \cite{deo2018convolutional} trys to predict. \underline{Black boxes} are surrounding cars. \underline{Black-solid lines} are the observed history, \underline{red-dashed lines} are the ground truth in the future, \underline{yellow-dashed lines} are the predicted results (5 seconds) of our GRIP++ (GRIP has a similiar performance on this dataset), and the \underline{green-dashed lines} are the predicted results (5 seconds) of CS-LSTM \cite{deo2018convolutional}. Region from $-90$ to $90$ feet are observed areas.}
    \vspace{-4mm}
    \label{fig:visualized_result}
\end{figure*}

In Figure \ref{fig:visualized_result}, we visualize several prediction results in mild, moderate, and congested traffic conditions (from left to right) using the datasets NGSIM I-80 and US-101. After observing 3 seconds of history trajectories, our model predicts the trajectories over 5 seconds horizon in the future. From Figure \ref{fig:visualized_result}, one can notice that:

\begin{itemize}
\item 1. From Figure \ref{fig:vis_sub_a} to Figure \ref{fig:vis_sub_c}, it is obvious that green-dashed lines (CS-LSTM) are longer than yellow-dashed lines (ours) and farther from the red-dashed lines (ground truth). This proves that when feeding the same history trajectories (all objects in the scene) to models, our proposed GRIP++ makes a better prediction for the central object than CS-LSTM. 

\item 2. In Figure \ref{fig:vis_sub_b}, our model precisely predicts the trajectory of the top car even when it is going to change lane in the next 5 seconds. In addition, the car in the left lane is affected by the top car, and our model still successfully predict the trajectory for the car in the left lane. 

\item 3. Our proposed GRIP++ can predict all objects in the scene simultaneously, while CS-LSTM can only predict the one located in the middle. Especially, in Figure \ref{fig:vis_sub_e}, we show a prediction result in a scene that involves 15 cars. In this scene, although some cars move slowly (vehicles in the middle lane) while others move faster (cars in the right lane), our proposed GRIP++ model is able to predict their future trajectories correctly and simultaneously. 
\end{itemize}

Based on these observations from the visualized results, we can conclude that our proposed scheme, GRIP++, indeed improves the trajectory prediction performance compared to the existing methods. Even though Figure \ref{fig:visualized_result} only shows straight high way scenario, our approach equally works for curved roads.

\section{\textbf{Conclusion}}
\label{sec:conclusion}
In this paper, we propose GRIP++ for autonomous driving cars to predict object trajectories in the future. The proposed model uses a graph to represent the interaction among all close objects and employs an encoder-decoder GRU-based  model to make predictions. Unlike some existing solutions that only predict the future trajectory for a single traffic agent each time, GRIP++ is able to predict trajectories for all observed objects simultaneously. The experimental results on two well-known highway and one urban traffic scenario datasets show that our proposed model achieves much better prediction results than existing methods and run $21.7$ times faster than one state-of-the-art scheme. Compared to our previous work, GRIP \cite{itsc19}, GRIP++ achieves similar performance in highway scenarios 83\% improvement in urban scenarios. We also conduct extensive ablation studies to understand how different design choices affect the trajectory prediction accuracy. In the near future, we hope to integrate GRIP++ into a route planning module and combine it with a deep learning based perception module to further evaluate the overall performance of these two modules. 
Subsequently, we intend to run this integrated perception and navigation module in prototype robotic cars in a testbed.

\vspace{2mm}
\noindent {\bf Acknowledgement:} This work is partially supported by a Qualcomm gift, an NSF CNS award 1931867, and a GPU donated by NVIDIA.
\vspace{-3mm}

{%\small
\bibliographystyle{IEEEtran}
\bibliography{references}
}

\end{document}